\documentclass[letterpaper, 10 pt, conference]{ieeeconf}  

\IEEEoverridecommandlockouts                              

\usepackage{graphicx}
\usepackage{amsmath}
\usepackage{cite}
\usepackage{mathtools, amssymb}
\usepackage{varioref}
\labelformat{equation}{(#1)}
\usepackage{amsfonts}

\usepackage{color}
\usepackage{enumerate}

\usepackage{mysymbol}

\usepackage[export]{adjustbox}

\usepackage[dvipsnames]{xcolor}
\usepackage{tikz,pgfplots}
\usepackage{color} 
\usetikzlibrary{matrix} 
\usetikzlibrary{arrows} 
\usetikzlibrary{calc} 
\usetikzlibrary{shapes}
\pgfplotsset{compat=1.7}

\usepackage{standalone}
\usetikzlibrary{arrows}
\usetikzlibrary{positioning}
\usetikzlibrary{backgrounds, fit}

\usepackage{amsthm}
\theoremstyle{plain}
\newtheorem{thm}{Theorem}

\theoremstyle{definition}
\newtheorem{asmptn}{Assumption}
\newtheorem{remark}{Remark}
\newtheorem{example}{Example}

\renewcommand{\epsilon}{\varepsilon}

\usepackage{algorithm}
\usepackage{algpseudocode}
\usepackage{algorithmicx}

\usepackage{epstopdf}

\title{\LARGE 
	\bf
	Federated Reinforcement Learning at the Edge
}
\author{Konstantinos~Gatsis~
	\thanks{
		The author is with the Department of Engineering Science, University of Oxford, Parks Road, Oxford, OX1 3PJ, UK. Email: konstantinos.gatsis@eng.ox.ac.uk.}%
}

\begin{document}
	
	\maketitle
	\thispagestyle{empty}
	\pagestyle{empty}

\begin{abstract}
Modern cyber-physical architectures use data collected from systems at different physical locations to learn appropriate behaviors and adapt to uncertain environments. However, an important challenge arises as communication exchanges at the edge of networked systems are costly due to limited resources. This paper considers a setup where multiple agents need to communicate efficiently in order to jointly solve a reinforcement learning problem over time-series data collected in a distributed manner. This is posed as learning an approximate value function over a communication network. An algorithm for achieving communication efficiency is proposed, supported with theoretical guarantees, practical implementations, and numerical evaluations. The approach is based on the idea of communicating only when sufficiently informative data is collected.
\end{abstract}

\section{Introduction}

Recent years have seen a shift in cyber-physical system architectures from systems running in isolation or local networks to systems connected to the cloud for outsourcing data and computations. This has given rise to the realization that the resulting bandwidth and communication requirements can become a bottleneck, and as a result, computing at the edge of the architecture is explored. Learning and adapting to data from sensors, robots, or vehicles located at the edge enable new applications such as cloud robotics \cite{chinchali2021network}, connected autonomous transportation systems, and the Industrial Internet-of-Things. Toward this end, this paper introduces a new framework for performing distributed reinforcement learning from data collected locally at individual agents/robots at the edge.

The accelerating developments in machine learning and reinforcement learning have increased the interest of the control community in using data-driven techniques. Specifically in the area of control of distributed and multi-agent systems, very recent developments include algorithms for multi-agent reinforcement learning  \cite{lowe2017multi, qu2020scalable,fattahi2020efficient}, reinforcement learning over networks/graphs  \cite{zhang2018fully, zhang2019distributed, cassano2020multi, doan2021finite},
as well as the search for appropriate parameterizations for these problems \cite{tolstaya2020learning}. However  when dealing with distributed learning at the edge, there is also the need for communication efficiency, especially if agents have high dimensional time-series data and operate over resource-limited communication networks.

The bulk of research in the area of distributed and communication-efficient learning, also termed federated learning, is focused on static machine learning problems, such as classification.  To overcome the communication bottleneck of sending high dimensional data, the main idea is to send  gradients of the objective with respect to the parameters being learned, instead of the data itself. Approaches based on gradient quantization and non-periodic  updates~\cite{konevcny2016federated, chen2018lag, reisizadeh2019fedpaq}, allocation of wireless resources~\cite{yang2019scheduling, amiri2020updateISIT}, or approaches exploiting the informativeness of the data~\cite{ACC21_Gatsis,L4DC21_Gatsis} 
are being explored.
For the problem of communication-efficient reinforcement learning, only now initial approaches emerge, including non-periodic updates for policy gradient methods \cite{chen2021communication}, distributed peer-to-peer network architectures~\cite{ornia2021event}, specific problems such as multi-armed bandits~\cite{mitra2021robust}, 
or regret analysis of online distributed reinforcement learning \cite{agarwal2021communication}.

\begin{figure}[t!]
	\centering
	{\resizebox{\columnwidth}{!}{\input{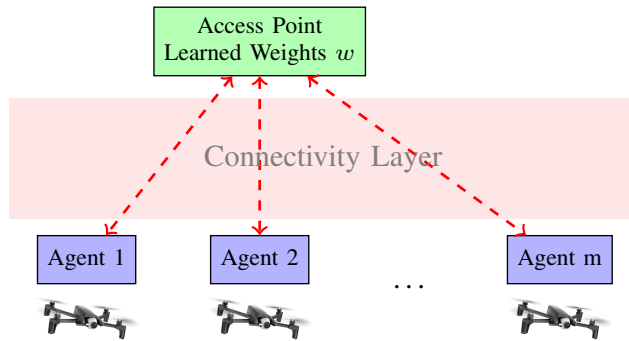}}	}
		\vspace{-5pt}
	\caption{Architecture for distributed reinforcement learning tasks over networks of agents. Agents are collecting state transition data and are communicating with an access point/server in order to learn a common model. Communication efficiency is achieved by communicating when data are more informative.
	}

	\label{fig:architecture}
	\vspace{-10pt}
\end{figure}

This paper introduces a new approach for communication-efficient reinforcement learning problems. To avoid costly exchange of time-series data, agents should assess \textit{how informative} are the data they collect, and communicate if the update will help the process of learning. The setup, described in Section~\ref{sec:setup} and shown in Fig.~\ref{fig:architecture}, involves multiple agents collecting state transitions and costs in a distributed manner for the purpose of learning approximate value functions, which is a central problem in reinforcement learning. Specifically linear value function approximation is considered. {
	While all recent approaches~\cite{chen2021communication, ornia2021event, mitra2021robust, agarwal2021communication} try to limit the amount of communication exchange  during reinforcement learning, the main novelty of this paper is that the communication cost is part of the performance criterion, and the resulting advantage is that algorithms which \textit{by design} efficiently tradeoff communication and learning are developed. 
} This is obtained both theoretically (Section \ref{sec:approach}) as well as in numerical examples in finite and continuous state spaces (Section~\ref{sec:numerical}).

The approach adapts ideas from the author's recent work on communication-efficient linear regression \cite{ACC21_Gatsis,L4DC21_Gatsis} to the reinforcement learning problem. On the technical side, the approach is different here as, to ensure convergence close to an optimal solution, the agents increasingly send less informative data as the number of iterations of the learning algorithm grows.
More broadly, the technical methodology relates to the problem of efficient control over networks \cite{EisenEtal19a,mamduhi2020cross}, event-triggered learning~\cite{solowjow2018event,ghosh2020eventgrad} and resource-aware optimization~\cite{vaquero2019convergence,magnusson2020maintaining, magnusson2017convergence, khirirat2021flexible}.

\section{Problem Setup}\label{sec:setup}

	The architecture examined in this paper, shown in Fig.~\ref{fig:architecture}, involves an access-point/server interested in solving a reinforcement learning task on data that are collected by multiple agents. Specifically, each agent is collecting independently state transitions (from a state $x^t$ to a new state $x^t_+$) and cost measurements (of the form $c^t$ at state $x^t$), and communicate with the server in order to jointly learn an approximate value function $V(x)$ parameterized by a vector of weights $w$.	The aim will be to achieve this with communication efficiency, i.e., without agents communicating all collected data all the time. An example scenario is presented, and the formal mathematical formulation follows next.
	
	\begin{example}[Motivating scenarios]
		As a first example, suppose the agents in  Fig.~\ref{fig:architecture} are robots/drones/vehicles with identical system dynamics. Then they can each collect data from their own trajectories (state transitions) as well as costs measuring how well they can achieve a common task, such as tracking a desired behavior. Transmitting all these data over the communication network would be costly.  A recent case study of this form by Google \cite{levine2018learning} considers the large scale training of control and manipulation policies using multiple robots collecting image data concurrently. As a second example, suppose the agents in  Fig.~\ref{fig:architecture} are sensors/actuators in an industrial setting collecting time-series data from multiple identical physical processes, for the purpose of improving the control of the processes.
	\end{example} 
	
	\subsection{Mathematical Problem Formulation}

	We assume a Markov Decision Process problem defined by the state space $X$, the action space $A$, the transition probabilities $\mathbb{P}(x_+\given x, a)$\footnote{or alternatively general state dynamics of the form $x_{+} = f(x, a, w)$ for some noise variable $w$}, and a cost function $c:X\times A \rightarrow \reals$ giving rise to costs of the form $c_k = c(x_k, a_k)$. We are interested in approximating the value function, or cost-to-go function of a given policy $a_k = \pi(x_k)$, defined as $V(x) := \mathbb{E} \left[ \sum_{k=0}^\infty \gamma^k c(x_k, a_k)\given x_0 = x \right] $ with a discount factor $\gamma\in(0,1)$. A general theoretical approach for finding the value function is to perform the Value Iteration Algorithm, which iteratively fixes a current guess of the value function $V^{\text{current}}(x), x\in X$, and updates it according to
	\begin{align}\label{eq:value_iteration}
		V^{\text{updated}} (x) &= c(x, \pi(x)) \notag\\
		&+ \gamma \, \mathbb{E} [V^{\text{current}}(x_+)\given x, a=\pi(x)]
	\end{align}
	for all points $x \in X$, and repeats the process again and again. Under technical conditions\cite[Vol. II, Ch. 2]{bertsekas}, this process can converge to the desired value function $V(x)$, which is a fixed point of \ref{eq:value_iteration} (Bellman equation). 
	
	In reinforcement learning and approximate dynamic programming, this iteration is performed approximately, using data collected by following the given policy. The data consists of multiple state transitions from states $x^t, t=0,1,2,\ldots$ to new states $x^t_+, t=0,1,2,\ldots$, following the above transition probabilities, and stage costs of the form $c^t$ at state $x^t, t=0,1,2,\ldots$. In general we can denote these as tuples $(x^t, c^t, x_+^t), t=0,1,2,\ldots$. In practice, these state transitions can be just segments from longer state trajectories. The state samples $x^t, t=0,1,2,\ldots$ are modeled as drawn from a distribution $d(x)$ in this paper.

	Moreover, instead of computing the updated value function in \ref{eq:value_iteration} in the space of all functions $\{V:X\rightarrow \reals\}$, a restricted function class is selected. In this paper we follow the commonly employed linear function class~\cite{sutton2018reinforcement, bertsekas}. In other words, we are trying to explain $V^{\text{updated}} (x)$ in \ref{eq:value_iteration} as a linear combination of basis functions (features) as
	\begin{equation}\label{eq:VFA}
		V^{\text{updated}} (x) \approx \sum_{i=1}^n w_i \phi_i(x) .
	\end{equation}
    Here $w \in \reals^n$ if the vector of weights (linear combination) to be learned, consisting of elements $w_i$, and $\phi_i(x)$ are fixed basis functions (not learned). Examples of general basis functions, especially when $X$ is a real vector space, include discretization functions, polynomial functions, radial basis functions, and others~\cite{sutton2018reinforcement, bertsekas}. 
    
    To pick one particular approximation in \ref{eq:VFA}, we are interested in minimizing the squared error
    \begin{equation}\label{eq:main_objective}
    	\underset{w \in \reals^n}{\text{minimize}} \; J(w) = \mathbb{E}_d \left[ V^{\text{updated}} (x) - \sum_{i=1}^n w_i \phi_i(x) \right]^2 
    \end{equation}
    where the expectation is with respect to a distribution $d(x)$, which corresponds to the distribution of the data collected by the agents. To sum up, instead of updating the value function according to \ref{eq:value_iteration}, we instead approximately update to the function $\sum_{i=1}^n w^*_i \phi_i(x)$ where $w^*$ is the vector solving \ref{eq:main_objective}.  After this approximation is found, the process \ref{eq:value_iteration} repeats. The current value function is reset and a new approximation is computed.
    
    A fundamental question then is how to solve problem \ref{eq:main_objective} from data. An approach is to perform what can be considered a stochastic gradient descent. Specifically, this in an iterative algorithm of the form
    \begin{equation}\label{eq:SGD}
    	w_{k+1} = w_k - \epsilon \hat{\nabla} J(w_k), \; k=0,1, \ldots
    \end{equation}
	where $\epsilon>0$ is a stepsize, and the gradient is approximated from the data tuples as
	\begin{align}\label{eq:stochastic_gradient}
		&\hat{\nabla} J(w_k) = \frac{1}{T} \sum_{t=0}^T \phi(x^t) \left(w_k^T \phi(x^t) - c^t - \gamma V^{\text{current}}(x_+^t)\right).
	\end{align}
	In this expression, $k$ refers to iterations of the algorithm \ref{eq:SGD}, while $t$ refers to identically distributed samples of state transitions and costs from the Markov Decision Process, and there are $T$ of them in total.
	It can be verified that this yields an unbiased estimate of the gradient of \ref{eq:main_objective}. After a large number of iterations $N$ of \ref{eq:SGD}, which also means after drawing many data samples, and with an appropriate stepsize explained below, the iterate $w_{N}$ will converge close to the optimal solution of the problem \ref{eq:main_objective} which we denote as $w^*$.

	\begin{remark}[Relation to other Reinforcement Learning Approaches]
		In this paper we illustrate how the above value function approximation in \ref{eq:main_objective} can be performed with communication efficiency. This approximation then needs to be performed at each iteration of the value iteration algorithm \ref{eq:value_iteration} -- see also Algorithm~\ref{algorithm}. The approach can also be extended to learn a Q-function approximation 
		but this is	not further discussed in this paper due to limited space. 
		More broadly, these algorithms are variants of Projected Value Iteration~\cite[Vol. II, Ch. 6]{bertsekas}, and are behind many successful reinforcement learning approaches~\cite{riedmiller2005neural, mnih2015human}. These algorithms are attractive because under technical conditions they converge to a unique point.
	\end{remark}

	\subsection{Communication-efficient reinforcement learning problem}
	
	Given the above modeling for a reinforcement learning task that needs to be solved, the communication problem is as follows. {At each iteration $k$, the server broadcasts the current weights $w_k$ to all agents.} Then each agent $i$ collects $T$ local data samples identically distributed (across samples and across agents), computes a local stochastic gradient $\hat{\nabla}_i J(w_k)$ from the available local data using formula \ref{eq:stochastic_gradient}, and decides whether to transmit this gradient update over the communication network to the receiving server. The server updates the current vector of weights $w_k$  depending on the information received from different agents. For simplicity of exposition the case of two agents is considered and theoretically analyzed, and experiments with more agents are conducted numerically in Section~\ref{sec:numerical}. This leads to the update rule at the server
	\begin{equation}\label{eq:dynamics_single_task}
		w_{k+1} = \left\{ \begin{array}{ll} w_k - \epsilon \hat{\nabla}_1 J(w_k) &\text{if  $1$ transmits}\\
			w_k - \epsilon \hat{\nabla}_2 J(w_k) &\text{if  $2$ transmits}\\
			w_k - \frac{\epsilon}{2} ( \hat{\nabla}_1 J(w_k) +\hat{\nabla}_2 J(w_k)) &\text{if both transmit}\\
			w_k &\text{if no transmits} \end{array}\right.
	\end{equation}
	We further denote with $\alpha_k^i \in \{1,0\}$ the decision for each agent $i=1,2$ to transmit or not. 
	
	{At the next iteration $k+1$ a new set of data is collected at each agent, a new stochastic gradient direction $\hat{\nabla}_i J(w_{k+1})$ with respect to the new vector $w_{k+1}$ is computed at each agent, and the process repeats until a final iteration $N$. The algorithm is also described in Algorithm~\ref{algorithm}.
		
		The aim will be to\textit{ avoid sending updates all the time in order to limit the communication burden}.} Hence we consider the average communication cost during all iterations and across all agents as
	\begin{equation}\label{eq:communication_cost}
		\frac{1}{N}\sum_{k=0}^{N-1}\, \frac{\alpha_k^1 + \alpha_k^2}{2} 
	\end{equation}
	On the other hand, we want to understand the progress of learning, hence we also measure how well the final set of weights $w_N$ solves \ref{eq:main_objective}, in other words we measure the cost function $J(w_N)$.

	We note that since the data points at each iteration and at each agent are random, so are the constructed stochastic gradient directions, and so are the vectors $w_{k}$, and so are the decisions $\alpha_k^i$ of the agents to transmit. We propose then to measure the efficiency of the implementation \textit{on average over the data points collected}. As a result, we establish the performance metric
	\begin{equation}\label{eq:performance_metric}
		\mathbb{E}_{\text{data}} \left[ \lambda \sum_{k=0}^{N-1}\, \frac{\alpha_k^1 + \alpha_k^2}{2N} + J(w_N) \right]
	\end{equation}
	Here $\lambda>0$ is a tuning parameter that is used to either penalize communication or learning performance.
	It is worth emphasizing then that there are two expectations in this paper. One is the integral defined in the objective $J(w)$ in \ref{eq:main_objective}, and another is the expected performance of the implementation in \ref{eq:performance_metric} computed as an integral over tuples $x^t, c^t, x_+^{t+1}$ -- in total $2NT$ tuples over agents, iterations, and samples -- and denoted as $\mathbb{E}_{\text{data}}$ to clarify the difference.

	\section{Theoretical scheme for communication-efficient reinforcement learning}\label{sec:approach}

	\begin{algorithm}[t!]
		\caption{Distributed Approximate Value Iteration Algorithm }
		\label{algorithm}
		\begin{algorithmic}[1]
			\State Fix a policy $\pi$
			\State Fix current value function guess $V^{\text{current}}(x)$.
			\State Fix initial weights $w_0$ at server \label{step:restart}
			\For{ Iteration $k=0,1,2, \ldots, N-1$}
			\State Server transmits weights $w_k$ to agents \label{line:begin}
			\State At each agent $i$ collect $T$ state transition samples following policy $\pi$ 
			\State At each agent $i$ compute stochastic gradient $\hat{\nabla}_i J(w)$ 
			\State At each agent $i$ implement communication \ref{eq:single_scheduling} with the approximation \ref{eq:approximate_gain_1} 
			\State At the server, update weights according to \ref{eq:dynamics_single_task} \label{line:end}
			\EndFor
			\State Set $V^{\text{updated}} (x) \leftarrow \sum_{i=1}^n w_{N}(i) \phi_i(x)$ using the final weights $w_N$
			\State Replace $V^{\text{current}}(x) \leftarrow V^{\text{updated}} (x) $ and go to Step \ref{step:restart}
		\end{algorithmic}
	\end{algorithm}

	{ The approach is based on the notion of performance gain which can be thought as a measure of how informative are the data collected at each agent at each time step with respect to the reinforcement learning problem. The gain at agent $i=1,2$ can be calculated by	
		measuring how much will the objective change if the agent sends the update.
		Whether this gain is negative or positive depends on the random direction of the update. 
		The proposed approach then is to send a gradient update if the gain is large enough. Mathematically we write
		\begin{equation}\label{eq:single_scheduling}
		\alpha_k^i = \left\{ \begin{array}{ll} 1 &\text{if } J(w_k - \epsilon \hat{\nabla}_i J(w_k)) - J(w_k) \leq -\frac{\lambda}{\rho^{N-1-k}} \\
		0 &\text{otherwise}\end{array}\right.
		\end{equation}
		where the scalar parameter $\lambda>0$ is the one defined in the performance criterion, and $\rho \in (0,1)$ is a parameter -- whose impact is further discussed in the theoretical results below.
		Intuitively this approach saves up communication resources, because the updates will be infrequent. Moreover, the term at the right hand side, which measures how informative are the data at the current iteration, is decreasing (in absolute value) as the number of iterations grow. Hence, at the beginning only very informative updates are transmitted, while as learning progresses, less informative updates are transmitted as well.

		\begin{asmptn}\label{as:matrix}
			The $n \times n$ symmetric matrix $\mathbb{E}_d \phi(x) \phi(x)^T$ is positive definite.
		\end{asmptn}
		This assumption guarantees the solution to the main problem \ref{eq:main_objective} is unique.
		
		\begin{asmptn}\label{as:epsilon}
			The step size $\epsilon>0$ in \ref{eq:dynamics_single_task} satisfies 
			\begin{equation}\label{eq:as_epsilon}
				\left|1- 2\epsilon  \lambda_i(\mathbb{E}_d \phi(x) \phi(x)^T )\right|<1
			\end{equation} 
		for all eigenvalues of the matrix at the right hand side.
		\end{asmptn}
		This assumption guarantees the step size is small enough so that convergence, even without communication constraints, is satisfied. A sufficient condition is $\epsilon<2/\lambda_{\max}$ where $\lambda_{\max}$ is the largest eigenvalue of the above matrix. Assumption \ref{as:matrix} guarantees that \ref{eq:as_epsilon} can be met.
	
		\begin{asmptn}\label{as:rho}
			The parameter $\rho$ in \ref{eq:single_scheduling} satisfies 
			\begin{equation}
				\rho \geq \max_i (1- 2 \epsilon  \lambda_i(\mathbb{E}_d \phi(x) \phi(x)^T ))^2
			\end{equation}
		for all eigenvalues of the matrix at the right hand side.
		\end{asmptn}	
		This assumption states that  measuring the informativeness of the data should decrease at a sufficiently slow rate. Assumptions \ref{as:epsilon} and \ref{as:rho} together guarantee that $\rho<1$ is a possible choice. 
		
		The main theoretical result of this paper is established next. 
		
		\begin{thm}[Communication-Efficient Value Function Approximation]\label{thm:theorem_single}
			Consider the optimization problem defined in \ref{eq:main_objective}. Consider the update rule in \ref{eq:dynamics_single_task}. Suppose $\hat{\nabla}_i J(w_k), i=1,2,$ are independent random variables with mean equal to $\nabla J(w_k)$ and covariance $G$ at each iteration $k$. Consider the communication strategy in \ref{eq:single_scheduling} with a fixed number of iterations $N$. Let Assumptions \ref{as:matrix}-\ref{as:rho} hold. Then we have that 
			\begin{align}\label{eq:main_result}
				&\mathbb{E}_{\text{data}} \left[ \lambda \sum_{k=0}^{N-1} \frac{\alpha_k^1+\alpha_k^2}{2N} +  J(w_{N}) \right] \leq 
				\lambda + J(w^*)  \nonumber\\
				&+ \rho^{N} [J(w_0) -J(w^*) ] + \frac{1- \rho^{N}}{1-\rho} \epsilon^2 \text{Tr}(\mathbb{E}_d \phi(x) \phi(x)^T  G) 
			\end{align}
			where $w^*$ is the optimal solution of \ref{eq:main_objective} and the expectation is with respect to the data collected until iteration $N$. 
		\end{thm}

	We have the following observations. The terms on the right hand side of \ref{eq:main_result} measure the suboptimality of learning, including the terms that have to do with the poor initialization $J(w_0) -J(w^*) $ and the noise of the gradients captured by the matrix $G$. As the number of iterations $N$ grows, the impact of the former is diminished, and only the latter remains. This expression also suggests that ideally one would pick the parameter $\rho$ to be the minimum allowed by the above Assumption \ref{as:rho}. 
	
	The theorem characterizes the tradeoff between communication and learning. By increasing the parameter $\lambda$, there is a higher penalty for communicating, and as a result, the agents communicate less often --- see following remark. As a consequence, learning performance will be impacted. But the above theorem \textit{guarantees by design} that there will be a graceful compromise between communication performance and learning -- the left hand side \ref{eq:main_result} cannot be arbitrarily poor.

	It is worth noting that, essentially, from the right hand side of \ref{eq:main_result}, the theorem states that the aggregate performance will be better than just having one agent only and transmitting all the time. It is possible to extend the analysis to get  bounds sharper than \ref{eq:main_result}, which will be the topic of future work.

	\begin{remark}
		In Theorem~\ref{thm:theorem_single} we assumed for simplicity that the stochastic gradients have bounded covariances that are constant over time. In reality for the problem above the covariance of the stochastic gradient in \ref{eq:stochastic_gradient} will depend on the current iterate $w_k$, but our choice can be justified. For example we can add a projection to a bounded set $\|w_k\|<M$ in the algorithm, so we only search over this restricted set of weights, resulting in bounded gradient noise covariances.  
	\end{remark}

	{ 
	\remark[Analysis over mutiple iterations]{The above theoretical analysis is performed for one iteration of Alg.~\ref{algorithm} (lines \ref{line:begin}-\ref{line:end}). As a result, at each iteration communication-efficiency is guaranteed. 
		After many iterations, the algorithm will converge to a neighborhood of the desired set of weights $w$, i.e., the value function approximation. This analysis will be explored in future work.
	}
}

	\section{Practical scheme for communication-efficient reinforcement learning}\label{sec:practical_approach}

	Despite the above guarantee, implementing the theoretical communication scheme in \ref{eq:single_scheduling} would be practically impossible because it requires information that is not known. Specifically it would require for every agent to  know the Markov Decision Process and the data distribution in order to compute the actual performance gain. Since these are unknown, one approach is to \textit{estimate the performance gain from the data}. In particular, since the objective function is quadratic, we can write the performance gain as
	\begin{align}\label{eq:gain_1}
		&J\left(w_{k}- \epsilon \hat{\nabla}_i J(w_k)\right) - J(w_k) = -\epsilon \hat{\nabla}_i J(w_k)^T \nabla J(w_k)
		\notag\\
		 & + \frac{1}{2} \epsilon^2 \hat{\nabla}_i J(w_k)^T \nabla^2 J(w_k) \hat{\nabla}_i J(w_k)
	\end{align}
	This is a quadratic function of the stochastic gradient $\hat{\nabla}_i J(w_k)$. 
Then we can approximate the gradient and the Hessian at each agent as
	\begin{align}
		&\nabla J(w_k) \approx \hat{\nabla}_i J(w_k)  \notag\\
		& \nabla^2 J(w_k) \approx \frac{1}{T} \sum_{t=0}^T \phi(x^t) \phi(x^t)^T
	\end{align}
	Hence, using the expression for the information gain in \ref{eq:gain_1}, we can approximate the gain at agent $i=1,2$ as\footnote{Overall at each agent these computations require $O(Tn)$ operations hence are scalable.}
	\begin{align}\label{eq:approximate_gain_1}
		&J\left(w_{k}- \epsilon \hat{\nabla}_i J(w_k)\right) - J(w_k) \approx \notag\\
		& -\hat{\nabla}_i J(w_k)^T\left[ I - \epsilon \frac{1}{2}  \frac{1}{T} \sum_{t=0}^T \phi(x^t) \phi(x^t)^T \right] \hat{\nabla}_i J(w_k)^T
	\end{align}
	It is important to emphasize that {this is no longer a simple quadratic function} of the stochastic gradient but a more complicated function - we note that the data appear both in the stochastic gradients as well as in the matrix in the middle. This approximate value of the gain may take again positive or negative values but it induces an approximation error/bias.
	
	As a result, we can implement the communication decision in \ref{eq:single_scheduling} with the approximation in \ref{eq:approximate_gain_1}. In this case we no longer have the performance guarantee in Theorem~\ref{thm:theorem_single}. In numerical evaluations however we see that despite the bias this mechanism performs very well.

}

{\begin{remark}Another intuitive approach to evaluate the informativeness of the data is to assume that the performance gain \ref{eq:gain_1} is large when the stochastic gradient has a large norm $\|\hat{\nabla}_i J(w_k)\|$ at an agent. However, recent work~\cite{ACC21_Gatsis,L4DC21_Gatsis} has demonstrated that this approach is not necessarily communication-efficient.
A different perspective is followed by \cite{chen2021communication}. When agents do not update their gradients at the server, the server keeps a memory of past received gradients and uses them for gradient descent in \ref{eq:dynamics_single_task}. The advantage of the present approach is that it introduces an explicit communication-learning tradeoff that can be controlled by the parameter $\lambda$. Technically, the approach \cite{chen2021communication} is developed for policy gradient methods, while here the approach is developed for value-function-based approaches.
\end{remark}}

\section{Numerical results}\label{sec:numerical}

\begin{figure}[t!]
	\centering
	\begin{tabular}{cc}
	{\resizebox{0.35\columnwidth}{!}{
			\tikzset{every loop/.style={min distance=20pt,in=90,out=60,looseness=5}}
			\begin{tikzpicture}
				\draw[step=1cm,gray,very thin] (0,0) grid (3,3);

				\node[] (drone)  at (1.5,2.5){\includegraphics[width=40pt]{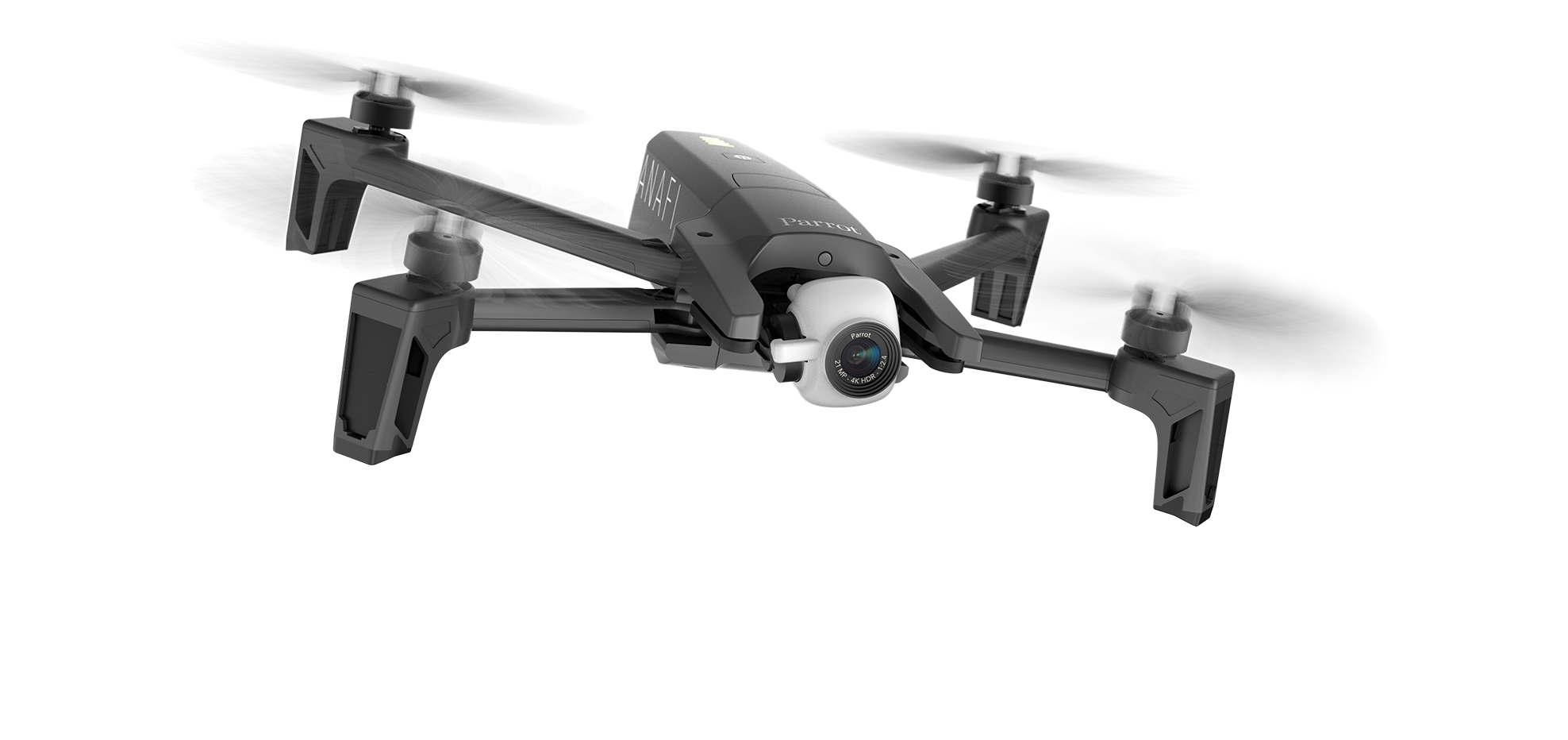}};
				
				\node[] (goal)  at (2.5,2.5){G};
				
				\fill [orange, opacity=0.2] (2,2) rectangle (3,3);
				
				
				
				\draw[white] (0,-2) rectangle (3,-1);
			\end{tikzpicture}
	}}
	%
	& \includegraphics[width=0.64\columnwidth]{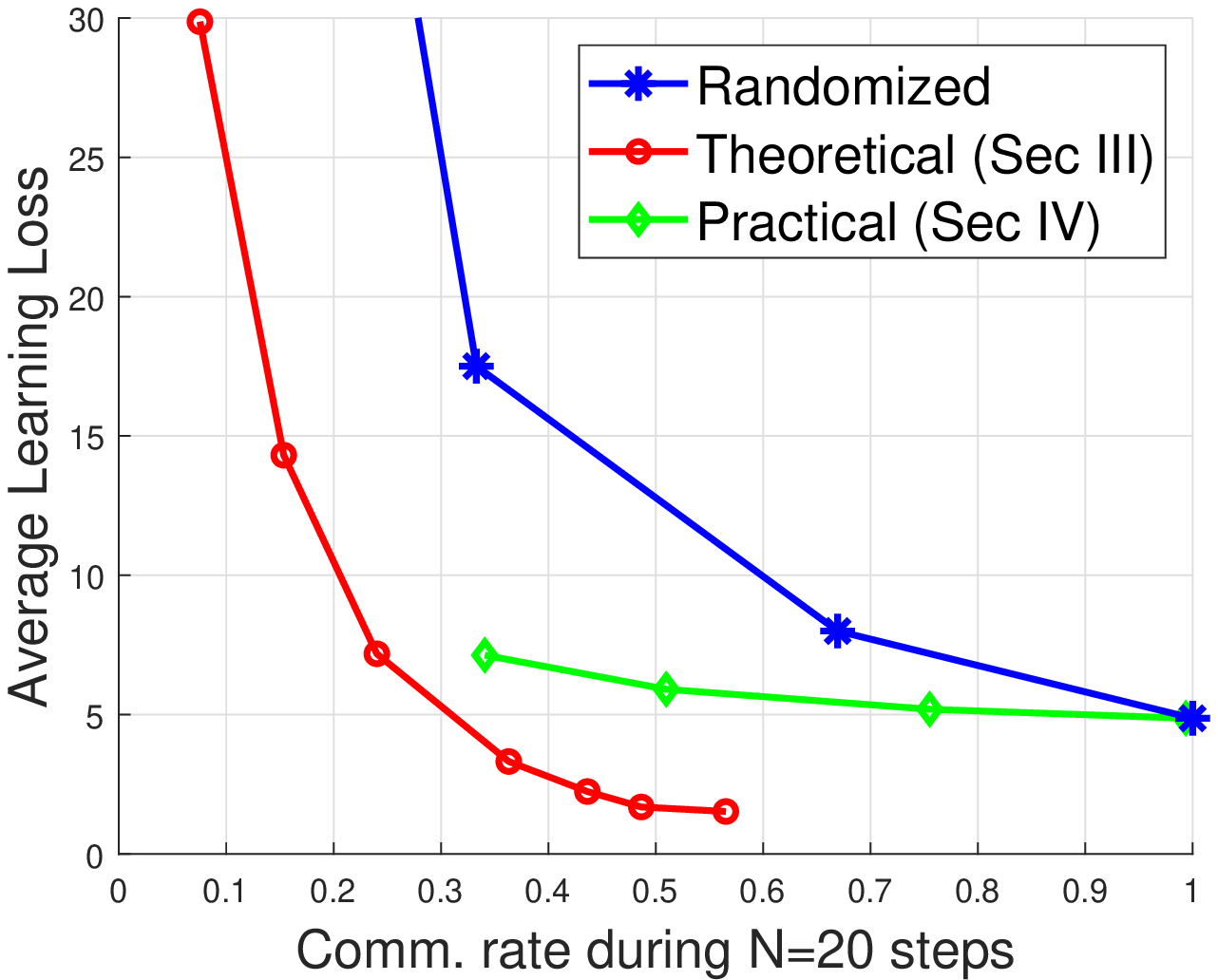}
	\end{tabular}
	\caption{Left: Grid example. Agents count the time it takes to reach the goal G while there is 50\% uncertainty in transitioning to the right at top row. Right: {Comparison between our communication approach in \ref{eq:single_scheduling} requiring the model to compute the performance gains versus estimating the gains by \ref{eq:approximate_gain_1}.} }
	\label{fig:tradeoff}
	\vspace{0pt}
\end{figure}

{We consider first a grid exploration example (Fig.~\ref{fig:tradeoff}) which is a finite state Markov Decision Process with state space $X$. 
	An agent can move in four directions subject to the boundary constraints, while at the top part of the grid there is a 50\% uncertainty in transitions to the right because of disturbances. The objective function is the total undiscounted ($\gamma=1$) time it takes to reach a desired goal location G. We collect measurements from multiple agents following the policy that randomizes over all actions at each state.

We perform one iteration of Algorithm~\ref{algorithm} following \ref{eq:value_iteration} where the initial value function is chosen randomly. In order to learn the exact value function we take the basis functions/features to be the indicators $\phi(1) = [1, 0, \ldots, 0]^T, \phi(2) = [0, 1, 0, \ldots, 0]^T$, etc in \ref{eq:VFA}.  We suppose each agent has few data tuples $T=10$ in every iteration, as described in Section~\ref{sec:setup}, chosen from a uniform distribution $d$ on this finite state space, and we take the stepsize to be $\epsilon=1$. We implement the theoretical approach \ref{eq:single_scheduling} and the practical approach using \ref{eq:approximate_gain_1}. We pick the parameter $\rho$ to be close to its smallest value allowed by Assumption \ref{as:rho}. For varying values of the parameter $\lambda$ we empirically compute the average communication rate \ref{eq:communication_cost}, and the average final learning performance $J(w_N)$ as defined in \ref{eq:main_objective}. The achieved tradeoff between the two is shown in Fig.~\ref{fig:tradeoff}. As a comparison, we also include the results when agents randomly decide whether the transmit their gradients or not, leading to an inefficient outcome.
}

For the theoretical algorithm \ref{eq:single_scheduling}, we observe that a very high communication efficiency can be achieved, i.e., by transmitting only a fraction of the time it is possible to find very good solutions to the value function approximation problem, hence achieving a good communication-learning tradeoff. For the practical approach using \ref{eq:approximate_gain_1} of course the learning loss is higher due to the bias introduced, but we also clearly observe a very good tradeoff between all agents communicating all the time versus scarce communications only when necessary.

\begin{figure*}[t!]
	\includegraphics[width=0.66\columnwidth]{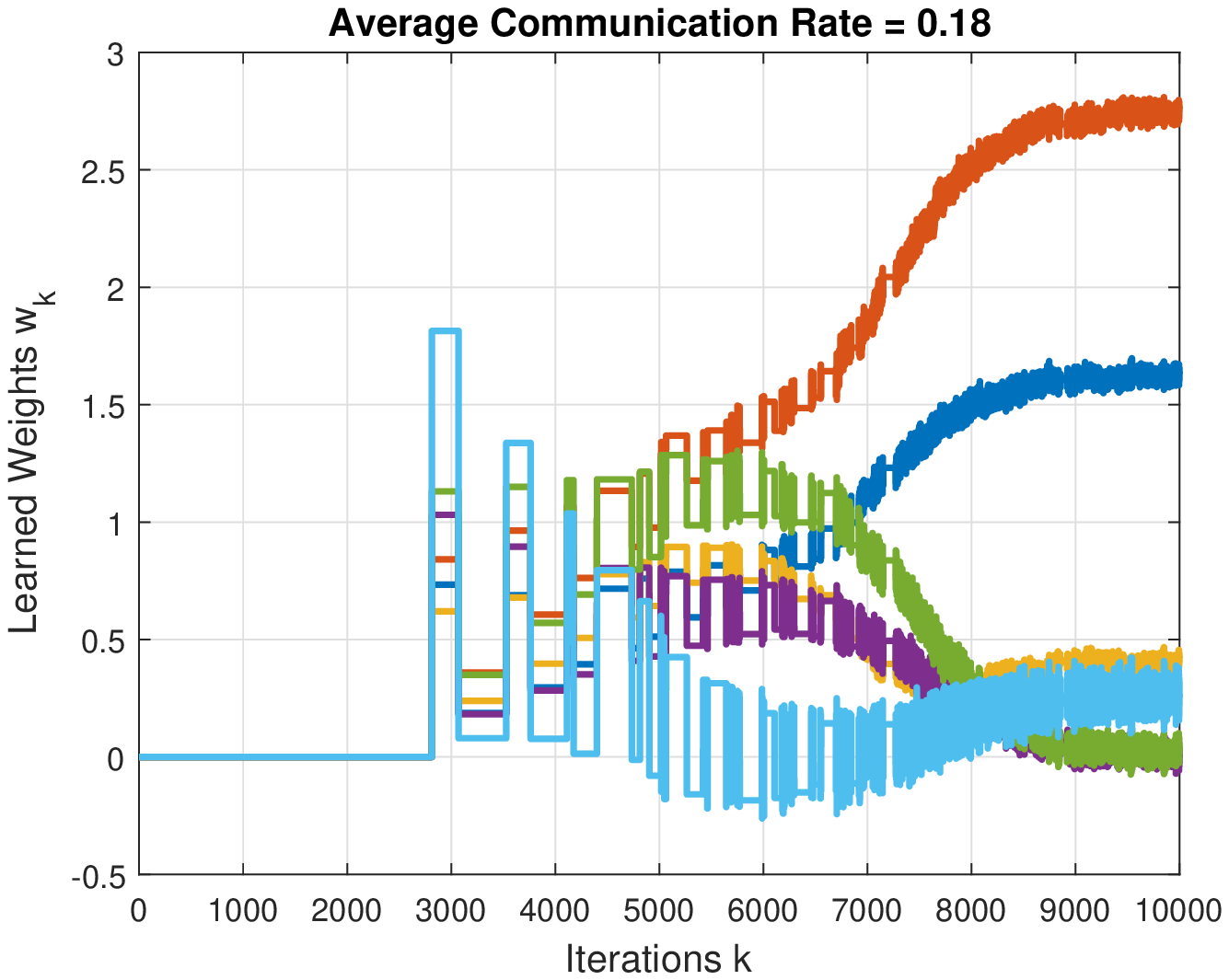}
	\includegraphics[width=0.66\columnwidth]{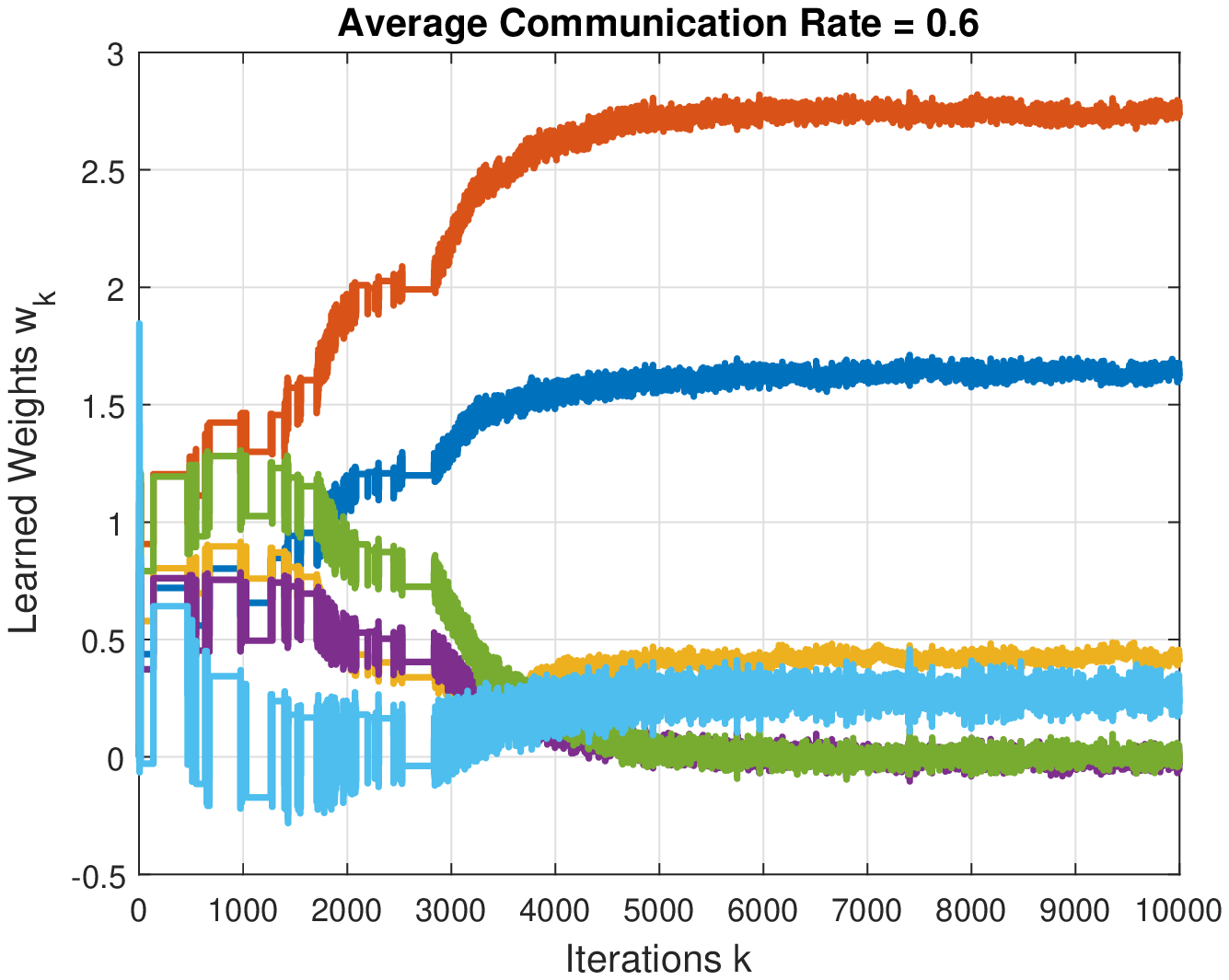}
	\includegraphics[width=0.66\columnwidth]{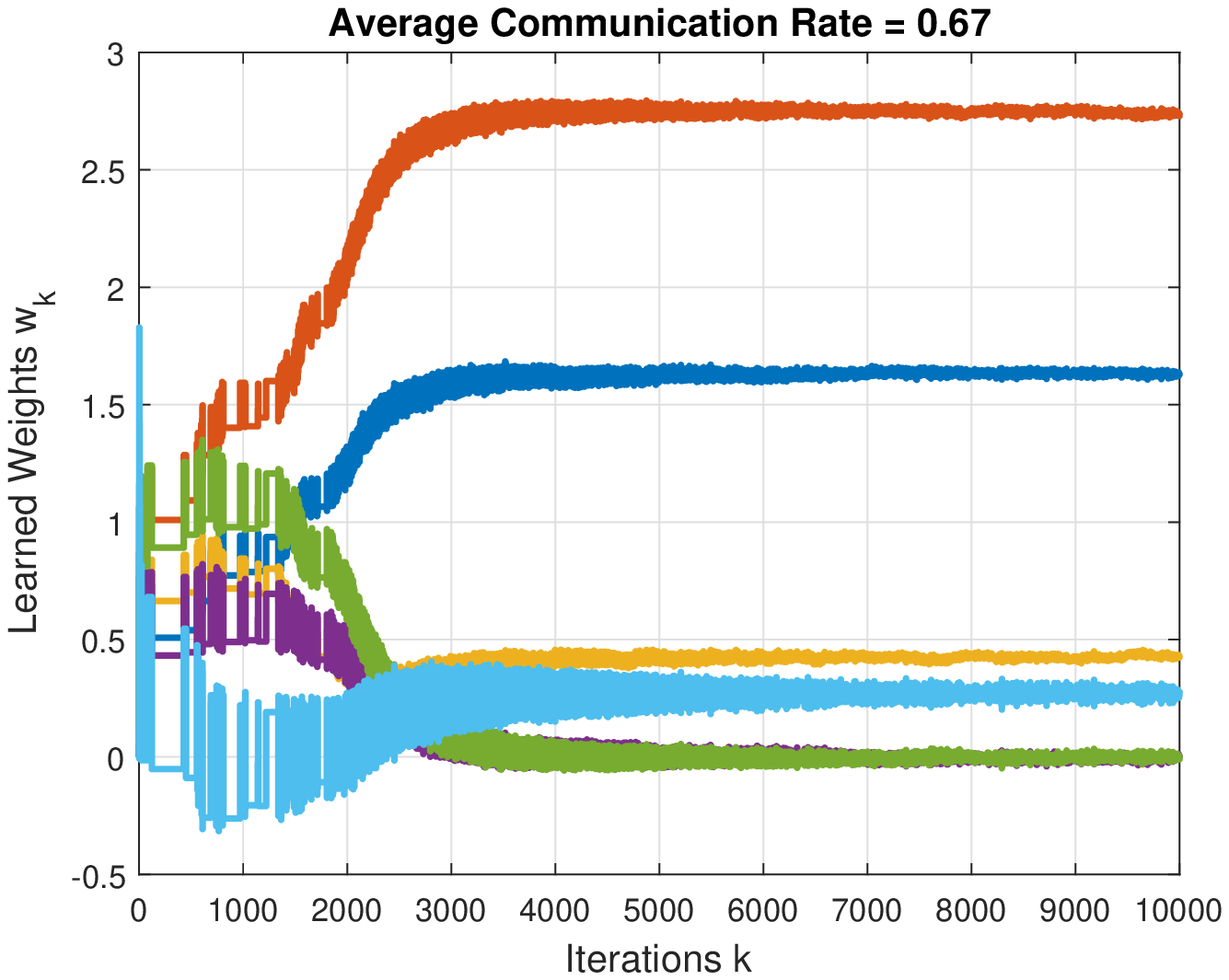}
	\caption{Left: {Communication-Efficient Distributed Value Function Approximation. Infrequent communication example is shown.} Middle: More frequent communication example  is shown. Right: A larger number of agents learns faster.}
	\label{fig:lambda_large}
	\vspace{0pt}
\end{figure*}

Then for a more complicated scenario we consider a continuous (uncountable) state space problem $X=\reals^2$. State transitions are given by a stochastic linear dynamical system $x_+= A x + w$, { with 
$A = \left[ \begin{array}{cc} 0.8 &-0.2\\ 0.1 &1 \end{array}\right]$, } 
where $w$ is a zero-mean Gaussian  noise variable with covariance $0.1$. We also consider a quadratic cost $c(x) = \|x\|^2$ and a discount factor $\gamma=0.9$. 
{We want to perform one value iteration following \ref{eq:value_iteration} where the initial value function is chosen randomly. We would like to approximate the updated value function per \ref{eq:VFA} as a linear combination of polynomial functions of the states with maximum degree 2 -- that is $\phi(x) = [x_1^2, x_2^2, x_1x_2, x_1, x_2, 1]^T \in \reals^6$ -- and we are searching for a vector of weights $w \in \reals^6$.}
We suppose each agent has at every iteration $T=10^3$ data tuples, as described in Section~\ref{sec:setup}, chosen from a uniform distribution $d$ on the space $[0,1]^2$. We take the stepsize to be $\epsilon=1$ and the parameter $\rho=0.999$. As expected when learning value functions on continuous state space, in this case we need many more iterations and a larger amount of data, in the order of $10^8$ data tuples.

We implement the practical communication algorithm with \ref{eq:approximate_gain_1}.
Fig.~\ref{fig:lambda_large} shows the evolution of weights for a large penalty $\lambda$  on communication rate. The figure shows that at the beginning no communication takes place because not very high informative gradients are found, while more communication takes place as learning progresses. 
Fig.~\ref{fig:lambda_large} shows the same setup with a smaller penalty $\lambda$ on communication, and we observe higher communication rate and faster learning of the weights. These observations verify the findings of Theorem~\ref{thm:theorem_single}.
Finally Fig.~\ref{fig:lambda_large} shows the same setup but with a larger number of agents,  10 instead of 2. We observe that learning happens faster, with almost the same amount of average communication rate.  This observation is not yet studied theoretically here and  will be explored in future work.

\section{Concluding remarks}

This paper examines the problem of solving reinforcement learning tasks over a network. To exploit the informativeness of the data, the notion of performance gain is explored and is shown numerically how this can be approximated from the data without further model knowledge. The approach is supported theoretically and numerically. 
Ongoing work explores the use of the approach in more complex networks and reinforcement learning algorithms.

\appendix

Within this proof for brevity we denote the stochastic gradients and the performance gains as
\begin{equation}
	g_k^i := \hat{\nabla}_i J(w_k), \qquad \lambda_k := \frac{\lambda}{\rho^{N-k-1} N}
\end{equation}
respectively.

\subsection{Proof of Theorem~\ref{thm:theorem_single}}

\begin{proof}
	
	Note that by the dynamics in \ref{eq:dynamics_single_task} we can write
	\begin{align}\label{eq:four_cases}
		J(w_{k+1}) &= (1- \alpha_k^1)(1- \alpha_k^2) J(w_k) \notag\\
		&+ \alpha_k^1 (1- \alpha_k^2) J(w_k - \epsilon g_k^1)
		\notag \\ 
		&+ (1- \alpha_k^1)\alpha_k^2  J(w_k - \epsilon g_k^2) \notag\\
		&+ \alpha_k^1 \alpha_k^2 J(w_k - \epsilon/2 g_k^1- \epsilon/2 g_k^2), 
	\end{align}
	depending on each of the four cases.
	Then due to the convexity of the problem we have for the last case the bound 
	\begin{equation}
		J(w_k - \epsilon/2 g_k^1- \epsilon/2 g_k^2) \leq 1/2 J(w_k - \epsilon g_k^1) + 1/2 J(w_k - \epsilon g_k^2).
	\end{equation}
	Substituting this bound in \ref{eq:four_cases} and after a  rearrangement of terms we get
	\begin{align}\label{eq:more_cases}
		J(w_{k+1}) &\leq \frac{1}{2} (1- \alpha_k^2) \Big[ (1- \alpha_k^1)J(w_k) + \alpha_k^1 J(w_k - \epsilon g_k^1) \Big] \notag\\
		&+ \frac{1}{2}\alpha_k^1 J(w_k - \epsilon g_k^1) 
		\notag \\ 
		&+ \frac{1}{2} (1- \alpha_k^1) \Big[ (1- \alpha_k^2)J(w_k) + \alpha_k^2 J(w_k - \epsilon g_k^2) \Big]
		\notag\\
		&+ \frac{1}{2} \alpha_k^2 J(w_k - \epsilon g_k^2)
	\end{align}
	Then we have for agent 1
	\begin{equation}
		\frac{1}{2} \lambda_k \alpha_k^1 \leq \frac{1}{2}  \lambda_k (1- \alpha_k^2)  \alpha_k^1 + \frac{1}{2}  \lambda_k \alpha_k^1
	\end{equation}
	and similarly for agent 2. Adding these with \ref{eq:more_cases}  we get
	\begin{align}\label{eq:more_more_cases}
		&\frac{1}{2} \lambda_k (\alpha_k^1+\alpha_k^2) + J(w_{k+1}) \notag\\
		&\leq \frac{1}{2} (1- \alpha_k^2) \Big[ \lambda_k \alpha_k^1  + (1- \alpha_k^1)J(w_k) + \alpha_k^1 J(w_k - \epsilon g_k^1) \Big] \notag\\
		&+ \frac{1}{2}\alpha_k^1 [\lambda_k + J(w_k - \epsilon g_k^1) ] \notag\\
		&+ \frac{1}{2} (1- \alpha_k^1) \Big[ \lambda_k \alpha_k^2  +  (1- \alpha_k^2)J(w_k) + \alpha_k^2 J(w_k - \epsilon g_k^2) \Big] \notag\\
		&+ \frac{1}{2}\alpha_k^2 [\lambda_k + J(w_k - \epsilon g_k^2) ] 
	\end{align}

	Then the terms in the brackets can be bounded. Note that due to the choice in \ref{eq:single_scheduling} the following inequality holds for all times (technically it holds almost surely as all the variables involved are random variables)
	\begin{equation}
		\lambda_k \alpha_k + (1-\alpha_k^i) J(w_k) + \alpha_k^i J (w_k - \epsilon g_k^i) \leq \lambda_k + J(w_k - \epsilon g_k^i).
	\end{equation}
	This can be easily verified by examining the two cases $\alpha_k^i =0$ or $1$ separately.
	Substituting this inequality for agents $i=1,2$ in \ref{eq:more_more_cases} we get
	\begin{align}
		&\frac{1}{2} \lambda_k (\alpha_k^1+\alpha_k^2) + J(w_{k+1}) \\
		&\leq \frac{1}{2} (1- \alpha_k^2) \Big[ \lambda_k + J(w_k - \epsilon g_k^1) \Big]
		+ \frac{1}{2}\alpha_k^1 [\lambda_k + J(w_k - \epsilon g_k^1)]
		\notag \\ &+ \frac{1}{2} (1- \alpha_k^1) \Big[ \lambda_k + J(w_k - \epsilon g_k^2) \Big]
		+ \frac{1}{2} \alpha_k^2 [\lambda_k + J(w_k - \epsilon g_k^2)] \notag
	\end{align}
	which after rearranging terms gives
	\begin{align}\label{eq:bound_more_cases}
		&\frac{1}{2} \lambda_k (\alpha_k^1+\alpha_k^2) + J(w_{k+1}) \\
		&\leq \lambda_k + \frac{1}{2} (1- \alpha_k^2) J(w_k - \epsilon g_k^1) 
		+ \frac{1}{2}\alpha_k^1  J(w_k - \epsilon g_k^1)
		\notag \\ &+ \frac{1}{2} (1- \alpha_k^1)  J(w_k - \epsilon g_k^2) 
		+ \frac{1}{2} \alpha_k^2  J(w_k - \epsilon g_k^2)
	\end{align}
	Taking expectation over the stochastic gradients $g_k^1$ and $g_k^2$, conditioned on the current iterate $w_k$, and using the symmetry of the problem with respect to agents $i=1,2$ we get that 
	\begin{align}\label{eq:intermediate}
		&\mathbb{E}[ \frac{1}{2} \lambda_k (\alpha_k^1+\alpha_k^2) + J(w_{k+1}) \given w_k] \leq \lambda_k +\\
		&
		\mathbb{E}[1 - \alpha_k^i\given w_k] \mathbb{E}[J(w_k - \epsilon g_k^i)\given w_k] + \mathbb{E}[\alpha_k^i J(w_k - \epsilon g_k^i)\given w_k] \notag
	\end{align}
	Then we have the following key fact, which {is shown separately in Appendix \ref{sec:appendix_technical}, }
	\begin{equation}\label{eq:main_comparison}
		\mathbb{E}[\alpha_k^i J(w_k - \epsilon g_k^i)\given w_k] \leq \mathbb{E}[\alpha_k^i\given w_k] \;  \mathbb{E}[J(w_k - \epsilon g_k^i)\given w_k]
	\end{equation}
	Substituting this bound in \ref{eq:intermediate}, we get
	\begin{align}\label{eq:descent_result}
		\mathbb{E}[ \frac{1}{2} \lambda_k (\alpha_k^1+\alpha_k^2) +  J(w_{k+1}) \given w_k] &\leq 
		\lambda_k + \mathbb{E}[J(w_k - \epsilon g_k^i)\given w_k] 
	\end{align}
	Then given the fact that the function $J(w)$ in \ref{eq:main_objective} is quadratic, i.e., can be written in the form 
	\begin{equation}
		J(w) = (w-w^*)^T\mathbb{E}_d \phi (w-w^*) + J(w^*)
	\end{equation}
	where we denoted $\Phi: = \mathbb{E}_d \phi(x) \phi(x)^T$. From the property of the stochastic gradient that the mean is unbiased $\mathbb{E}g_k^i= \nabla_{w} J(w_k) = 2\phi (w-w^*) $ with a constant variance $G$, we get that
	\begin{align}
		&\mathbb{E}[J(w_k - \epsilon g_k^i)\given w_k] = \notag\\
		&\mathbb{E}[(w_k - \epsilon g_k^i - w^*) \Phi (w_k - \epsilon g_k^i - w^*)] +J(w^*)\notag\\
		& =  (w_k-w^*)^T (I - 2 \epsilon \Phi)^T \Phi(I - 2 \epsilon \Phi) (w_k-w^*) \notag\\
		& + \epsilon^2 \text{Tr}(\Phi  G) +  J(w^*)
	\end{align}
	Then by Assumption \ref{as:rho} we can bound 
	\begin{align}
		\mathbb{E}[ J(w_k - \epsilon g_k)&\given w_k] \leq \rho J(w_k) + \epsilon^2 \text{Tr}(\Phi  G) + (1-\rho)J(w^*)
	\end{align}
	Substituting this in \ref{eq:descent_result} we get,
	\begin{align}
		&\mathbb{E}[ \lambda_k \frac{\alpha_k^1+\alpha_k^2}{2} +  J(w_{k+1}) \given w_k] \leq 
		\lambda_k + \rho J(w_k) \\
		&+ \epsilon^2 \text{Tr}(\Phi  G) + (1-\rho)J(w^*) 
	\end{align}
	Then for time-varying parameter $\lambda_k = \frac{\lambda}{\rho^{N-k-1} N}$ we get
	\begin{align}
		&\mathbb{E}[ \lambda \frac{\alpha_k^1+\alpha_k^2}{2N} +  \rho^{N-k-1}J(w_{k+1}) \given w_k] \leq 
		\frac{\lambda}{N} + \rho^{N-k} J(w_k) \\
		&+ \rho^{N-k-1} \epsilon^2 \text{Tr}(\Sigma_x  G) + \rho^{N-k-1} (1-\rho)J(w^*) 
	\end{align}
	Taking expectation on both sides with respect to the variable $w_k$, iterating over time $k=0, \ldots, N-1$, summing up, and removing the canceling terms on both sides, we get
	\begin{align}
		&\mathbb{E}\left[ \lambda \sum_{k=0}^{N-1} \frac{\alpha_k^1+\alpha_k^2}{2N} +  J(w_{N}) \right] \leq 
		\lambda + \rho^{N} J(w_0)  \\
		&+ \frac{1- \rho^{N}}{1-\rho} \left[\epsilon^2 \text{Tr}(\Phi  G) +  (1-\rho)J(w^*)\right] 
	\end{align}
	and complete the desired result \ref{eq:main_result}.
\end{proof}

\subsection{Proof of \ref{eq:main_comparison}}\label{sec:appendix_technical}

Within this proof we drop the iteration index $k$.
Let us denote by $F(g)$ the distribution of the stochastic gradient $g := \hat{\nabla} J(w_k)$ at any agent. Then we can rewrite \ref{eq:main_comparison} as 
\begin{equation}
\int \alpha(g)J(w-\epsilon g) dF(g) \leq \int \alpha(g)dF(g) \int J(w-\epsilon g) dF(g)
\end{equation}
However, by definition of the communication rule \ref{eq:single_scheduling} we have that $\alpha(g)=1$ only when $J(w-\epsilon g) \leq J(w)-\lambda$ and zero otherwise. Let us define this set of values $S= \{g\in \reals^n: J(w-\epsilon g) \leq J(w)-\lambda\}$, which is allowed to be an empty set too. Then \ref{eq:main_comparison} is equivalent to
\begin{align}
&\int_S J(w-\epsilon g) dF(g) \leq \notag\\
&\int_S dF(g) \left[ \int_S J(w-\epsilon g) dF(g) + \int_{S^c} J(w-\epsilon g) dF(g)\right]
\end{align}
%
%
%
which is 
equivalent to
\begin{align}\label{eq:equivalent_inequality}
\int_{S^c} dF(g) \int_S J(w-\epsilon g) dF(g) \leq \int_S dF(g) \int_{S^c} J(w-\epsilon g) dF(g)
\end{align}
We can bound the left hand side because we can bound $J(w-\epsilon g)$ point wise on the set $S$ as
\begin{align}\label{eq:bound_1}
\int_{S^c} dF(g) \int_S J(w-\epsilon g) dF(g) \leq \int_{S^c} dF(g) \int_S dF(g) \; [J(w)-\lambda]
\end{align}
Further we can bound the right hand side of \ref{eq:equivalent_inequality} as
\begin{align}\label{eq:bound_2}
\int_S dF(g) \int_{S^c} J(w-\epsilon g) dF(g) \geq \int_S dF(g) \int_{S^c} dF(g) \; [J(w)-\lambda]
\end{align}
Combining \ref{eq:bound_1} and \ref{eq:bound_2} we verify \ref{eq:equivalent_inequality} and conclude the proof.
%

\bibliographystyle{ieeetr}
\bibliography{federated_learning,distributed_rl}

\end{document}